\documentclass[10pt,twocolumn,letterpaper]{article}

\usepackage[pagenumbers]{cvpr} 

\usepackage{array,multirow}

\usepackage{graphicx}
\usepackage{amsmath}
\usepackage{amssymb}
\usepackage{booktabs}
\usepackage{array,multirow}
\usepackage{makecell}

\usepackage{stmaryrd}
\usepackage{comment}
\usepackage{amsmath,amssymb} 
\usepackage{color}
\usepackage{bbm}
\usepackage{romannum}
\usepackage{caption}
\usepackage{graphicx}
\usepackage{enumitem}
\usepackage{diagbox}

\newcommand{\cmk}{\checkmark}

\newcommand*\samethanks[1][\value{footnote}]{\footnotemark[#1]}
\usepackage[dvipsnames]{xcolor}

\definecolor{cvprblue}{rgb}{0.21,0.49,0.74}
\usepackage[pagebackref,breaklinks,colorlinks,citecolor=cvprblue]{hyperref}

\def\paperID{825} 
\def\confName{CVPR}
\def\confYear{2024}

\title{ControlUDA: \textit{Control}lable Diffusion-assisted \textit{U}nsupervised \textit{D}omain \textit{A}daptation \\for Cross-Weather Semantic Segmentation}

\author{Fengyi Shen\textsuperscript{1,2,3}, Li Zhou\textsuperscript{1}, Kagan Kucukaytekin\textsuperscript{1}, Ziyuan Liu\textsuperscript{2}, He Wang\textsuperscript{3}\thanks{corresponding author}, Alois Knoll\textsuperscript{1}\samethanks\\
\noindent\textsuperscript{1}Technical University of Munich, \textsuperscript{2}Huawei Munich Research Center, \textsuperscript{3}EPIC Lab, Peking University\\
{\tt\small \textsuperscript{1}\{first.last\}@tum.de,knoll@in.tum.de,\textsuperscript{2}\{first.last\}@huawei.com,\textsuperscript{3}hewang@pku.edu.cn}
}

\begin{document}
\maketitle
\begin{abstract}
Data generation is recognized as a potent strategy for unsupervised domain adaptation (UDA) pertaining semantic segmentation in adverse weathers. Nevertheless, these adverse weather scenarios encompass multiple possibilities, and high-fidelity data synthesis with controllable weather is under-researched in previous UDA works. The recent strides in large-scale text-to-image diffusion models (DM) have ushered in a novel avenue for research, enabling the generation of realistic images conditioned on semantic labels. This capability proves instrumental for cross-domain data synthesis from source to target domain owing to their shared label space. Thus, source domain labels can be paired with those generated pseudo target data for training UDA. However, from the UDA perspective, there exists several challenges for DM training: $(i)$ ground-truth labels from target domain are missing; $(ii)$ the prompt generator may produce vague or noisy descriptions of images from adverse weathers; $(iii)$ existing arts often struggle to well handle the complex scene structure and geometry of urban scenes when conditioned only on semantic labels. To tackle the above issues, we propose ControlUDA, a diffusion-assisted framework tailored for UDA segmentation under adverse weather conditions. It first leverages target prior from a pre-trained segmentor for tuning the DM, compensating the missing target domain labels; It also contains UDAControlNet, a condition-fused multi-scale and prompt-enhanced network targeted at high-fidelity data generation in adverse weathers. Training UDA with our generated data brings the model performances to a new milestone (72.0 mIoU) on the popular Cityscapes-to-ACDC benchmark for adverse weathers. Furthermore, ControlUDA helps to achieve good model generalizability on unseen data. 
\end{abstract}

\section{Introduction}
\label{sec:intro}

Urban scene segmentation in adverse weather conditions ~\cite{zhao2017pyramid,xie2021segformer,sakaridis2021acdc} presents significant challenges, primarily due to the safety concerns during data collection and the high costs associated with annotating images in degraded visibility. An important method to tackle this problem is unsupervised domain adaptation (UDA)~\cite{sakaridis2019guided,sakaridis2020map,hoyer2023mic}, adapting knowledge from the labelled clear to the unlabelled adverse weathers. 

Among UDA solutions, a widely adopted one is to synthesize data via generative models, \textit{e.g.}, GAN-based~\cite{zhu2017unpaired,zheng2020forkgan,liu2017unsupervised} translators to alter the style of an image from the source to the target domain and improve the model adaptability~\cite{sun2019see,romera2019bridging,sakaridis2019guided,sakaridis2020map}. However, this tends to be sub-optimal. {\bf First}, in real scenario, an important fact is that the adverse target domain usually comes with multiple weather and illumination possibilities, thus the generative model is supposed to flexibly synthesize controllable cross-weather data, which is underexplored in previous GAN-based UDA methods. {\bf Second}, GANs in prior arts of UDA are usually trained from scratch on smaller datasets, failing to ensure high-fidelity data generation. Driven by the recent success of large-scale text-to-image diffusion models (DM)~\cite{rombach2022high}, there is a sweeping research trend to employ such generative models for downstream tasks~\cite{lin2023magic3d,wu2023datasetdm,liu2023detector,tan2023diffss,gong2023prompting} and also by tuning~\cite{ruiz2023dreambooth,zhang2023adding} to fit specific tasks defined on customized datasets. On this basis, in this work, we exploit to devise a controllable generative method which enables high-fidelity multi-weather data generation to particularly assist UDA segmentation under this challenging scenario.


In the context of UDA segmentation in adverse weathers, due to the presence of technologies such as ControlNet~\cite{zhang2023adding}, coupled with the fact that a common label space is shared between source and target domains, an idea of synthesizing high-fidelity pseudo target domain data from source ground-truths arises. Subsequently, a segmentation model can be trained on this paired synthesized dataset to bridge the domain gap. Nonetheless, {\bf a fundamental paradox} here lies in the absence of target domain labels required to train such a DM. To address this inherent chicken-and-egg problem, we propose a solution that involves harnessing the target prior obtained from a pretrained UDA segmentor to tune the DM. This strategy allows for the generation of diverse pseudo target domain data from source domain labels post DM tuning. The synthesized data can then be employed to enhance UDA training, effectively transforming what initially appears to be an unsolvable problem into one amenable to label supervision.
Nevertheless, due to the scene complexity of driving data and their degraded visibility, even the most recent generative model, \textit{e.g.,} ControlNet~\cite{zhang2023adding}, faces several difficulties. {\bf First of all}, due to the scene complexity (e.g., object overlap, small objects) of driving data and their degraded visibility, existing methods such as ControlNet cannot ensure high quality image generation. {\bf Additionally}, if extra visual information has to be inserted to facilitate the training, according to~\cite{zhang2023adding}, a separate ControlNet has to be trained to support condition fusion, which is computationally impractical.  
{\bf More importantly}, the prompt generator (e.g., Blip~\cite{li2022blip}) used by DMs does not produce visual prompts with detailed and accurate class-wise semantics, meanwhile it fails in many cases to generalize well on data under adverse weathers, which brings noise to DM tuning as well as data generation since prompt belongs to one of the primary control signals.\\
 To address the aforementioned issues in ControlNet, we propose UDAControlNet which focuses on the following improved aspects: (\textbf{\textit{i}}). we prepare the input batch in multi-resolution setting to let the model observe both local and global information of the input condition, expecting to improve its capability of generating objects at various scales; (\textbf{\textit{ii}}). we introduce a residual condition fusion (RCF) module, allowing us to incorporate structure information while still prioritizing on the semantic condition. This not only resolves ControlNet's inflexibility of multi-condition fusion, but also differentiates the overlapped object instances in data generation.
(\textbf{\textit{iii}}). we augment the default Blip prompt with semantic information derived from the label condition for better semantic alignment and also specify the name of the incoming target sub-domain to improve weather controllability after training. Our enhanced prompt alleviates the impact when the Blip prompt is imperfect, resulting in better text-to-image diffusion for our targeted task.\\ 
Compared to previous UDA methods which adopt GANs, our UDAControlNet can flexibly synthesize higher fidelity and controllable pseudo target images conditioned on source labels, which, in return, are adopted to train a UDA segmentor for further performance boost. 
Induced by our UDAControlNet, a novel UDA training pipeline can be established, and in our work we refer to this as ControlUDA framework. ControlUDA pushes the performance of UDA segmentation in adverse weathers further towards its upper bound. To the best of our knowledge, this is the first trial of tuning large text-to-image generation models to particularly assist the task of UDA segmentation for urban scenes under adverse weathers.
Our contributions are below:
\begin{itemize}[noitemsep,nolistsep]
\item We propose to take advantage of target prior for large DM tuning, resulting in a target-specific generative model, whose generated pseudo target data conditioned on the source ground-truth labels can be, in consequence, supervised by the same labels to reinforce UDA training; 
\item On top of ControlNet, we propose a UDAControlNet which allows for adjustable multi-condition fusion of semantic and structure, together with our enhanced prompts to support weather-controllable and semantic-aligned DM tuning. It outperforms existing methods on data generation for adverse weathers; 
\item Our ControlUDA achieves consistent performance gain over prior arts on UDA segmentation under adverse weathers. 
\end{itemize}
%

\section{Related Work}
\label{sec:related_work}
{\bf Unsupervised Domain Adaptation (UDA).} For research under adverse conditions, due to the lack of training labels, there has been a rising interest in adapting knowledge trained from the labelled clear weather to unlabelled adverse weathers via UDA~\cite{dai2018dark,sakaridis2019guided,wu2021dannet,wu2021one,gao2022cross}. Self-training~\cite{wu2021one,hoyer2022daformer,xie2023sepico,gao2022cross,bruggemann2023contrastive,shen2023loopda} has been adopted by most existing methods since generating pseudo-labels for the target domain brings substantial performance gain to the model.
Other works~\cite{sakaridis2019guided,sakaridis2020map,wu2021dannet,wu2021one,shen2023loopda,bruggemann2023refign} utilize GPS information, taking clear reference images in a weakly supervised setting to provide extra guidance to improve pseudo supervision on target images. Without considering reference images, \cite{hoyer2022daformer,hoyer2022hrda,hoyer2023mic} introduces transformer-based architectures with affiliated training strategies and brings the UDA performances to a new milestone. Nevertheless, our argue that the limit of this task is far from being reached, and marrying UDA with large diffusion models can lift the upper bound of UDA segmentation in adverse conditions even without relying on extra reference data.\\
{\bf GAN-Based Style Transfer.} Enriching diversity by performing data augmentation~\cite{simard2003best,devries2017improved,devries2017dataset,zhang2017mixup,yun2019cutmix,cubuk2020randaugment,olsson2021classmix,kundu2021generalize,tranheden2021dacs,araslanov2021self,shen2023diga} has been proven to be a universal trick for most computer vision problems. Regarding domain adaptive semantic segmentation in adverse conditions, early attempts based on GANs~\cite{goodfellow2014generative,arjovsky2017wasserstein,mao2017least,jolicoeur2018relativistic_old,karras2019style} seek to transfer the style from source to target domain (e.g., daytime into nighttime style), and then optimize the model with the same source labels to improve its accuracy on the target domain~\cite{sun2019see,romera2019bridging,sakaridis2019guided,sakaridis2020map}. However, such domain style transfer usually show limited performance gain as source-specific information cannot be totally removed by the translated image once being exposed to the network. We tackle this issue by leveraging target prior labels for data generation training, which allows us to only generate target-specific features in the output images.\\
{\bf Diffusion Models (DM).} Denoising diffusion probabilistic models~\cite{ho2020denoising,nichol2021improved,song2020denoising,choi2021ilvr,dhariwal2021diffusion} have recently attracted much attention and have been applied to various vision tasks including semantic segmentation~\cite{brempong2022denoising,baranchuk2021label}. On that basis, stable diffusion (SD)~\cite{rombach2022high} has brought generative models to a new dimension owing to its high fidelity text-to-image generation capability. ~\cite{tan2023diffss,gong2023prompting,xu2023open,wu2023datasetdm} improve label-efficient semantic segmentation by utilizing either the intermediate representations or outputs of a pre-trained and fixed SD model. \cite{benigmim2023one} further points out that mining the rich hidden knowledge inside SD model by fine-tuning it on a customized target dataset can boost the segmentation performance. However, regarding data under adverse conditions, the default prompt generator will produce vague and incorrect descriptions of the input scene in many cases, which brings noisy features into the tuning process. Therefore, based on the recent advanced method ControlNet~\cite{zhang2023adding}, we devise a multi-scale structure-aware and prompt-enhanced diffusion tuning approach which is particularly targeted at high-quality image generation to boost UDA segmentation in adverse conditions.\\


\begin{figure*}[t!]
\centering
\includegraphics[width=2.0\columnwidth]{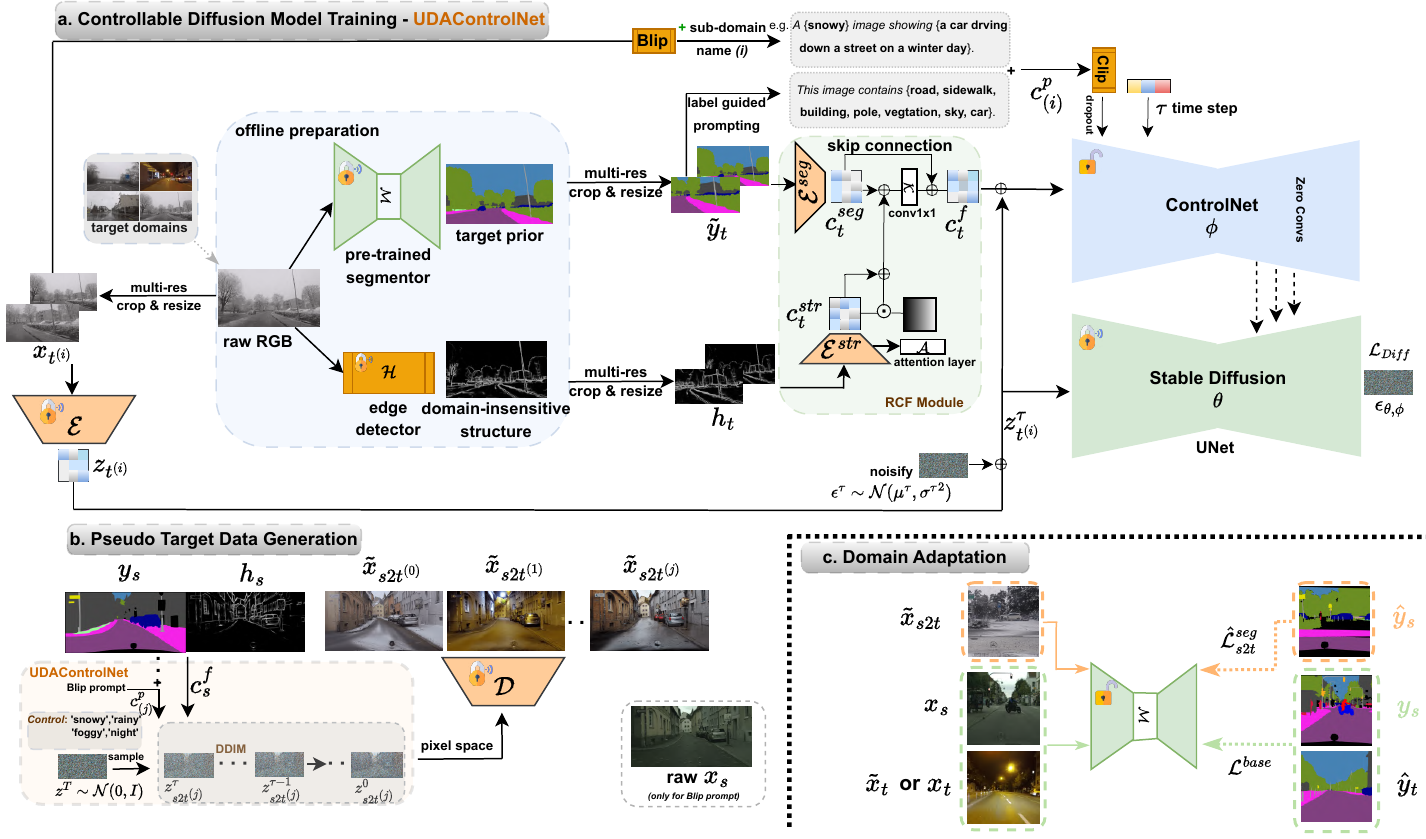}
\caption{{\bf An algorithmic overview of ControlUDA framework}. (a) depicts the training procedure of our UDAControlNet conditioned on prior knowledge from target domain, as described in Sec.~\ref{sec:training}; (b) demonstrates how data sampling can be performed with our trained UDAControlNet to synthesize various pseudo target data from a single source label (Sec.~\ref{sec:inference}); (c) shows how the performance of domain adaptive semantic segmentation in adverse weathers can be boosted via refinement training with our generated data (Sec.~\ref{sec:task}).}
\label{fig:diagram}
\end{figure*}

\section{Method}
\label{sec:method}
In this section we present ControlUDA (see Fig.~\ref{fig:diagram}), a novel framework that studies in what way tuning a large text-to-image diffusion model can be made beneficial to the task of unsupervised domain adaptive semantic segmentation in adverse weathers. Thus, we aim to build a bridge between these two research areas. To notate our research problem, let $(\mathcal{X}_s, \mathcal{Y}_s)$ denote the source domain clear weather dataset and $x_s \in \mathcal{X}_s$ is a source RGB image or mini-batch with semantic label map $y_s \in \mathcal{Y}_s$. 
And $\mathcal{X}_{t}$ denotes the target domain adverse weather dataset which consists of multiple target sub-domains $\mathcal{X}_{{t}^{(i)}}$, $i \in R$ indicating different challenging weather and illumination scenarios. $x_t \in \mathcal{X}_t$ represents an unlabelled training image or mini-batch from the target domain. $\theta$ denotes the pre-trained stable diffusion (SD) and $\phi$ is the ControlNet. The goal is to leverage the labelled $(\mathcal{X}_s, \mathcal{Y}_s)$ and the generative power of $\theta$ and $\phi$ for data synthesis, to obtain a segmentor $\mathcal{M}$ that performs accurate segmentation for $\mathcal{X}_t$ in adverse weathers as well.


\subsection{Controllable Diffusion Model Training}
\label{sec:training}
{\bf Acquisition of Target Prior Knowledge}
Due to the absence of $\mathcal{Y}_{t}$ in UDA training, to take the best advantage of existing source labels $\mathcal{Y}_{s}$, it is ideal to have another set of target domain images $\tilde{\mathcal{X}}_t$ which class-wisely and pixel-wisely share $\mathcal{Y}_{s}$ as their ground-truths (GT). The latest advancements in large-scale DMs for high-fidelity conditional image generation~\cite{rombach2022high,zhang2023adding} have paved the way for such research. For instance, an idea is inferring $\mathcal{Y}_{s}$ to a diffusion model trained on the target domain. However, the chicken and egg problem here is that training such a target-specific diffusion model requires $\mathcal{Y}_{t}$ to exist with $\mathcal{X}_t$, because a model trained with $(\mathcal{X}_s, \mathcal{Y}_s)$ has no target-awareness and thus cannot generate high-quality target data. Thanks to the recent progress of UDA~\cite{wu2021dannet,hoyer2022daformer,hoyer2023mic}, a trained segmentor is able to provide decent segmentation labels $\tilde{y}_t$ for $x_{t}$. Therefore, We argue that such target domain prior knowledge from a trained UDA segmentor can be a suitable settlement for the chicken and egg problem, even though the predictions are still more or less noisy. Here, our hypothesis is: if training the DM on $(\mathcal{X}_t, \tilde{\mathcal{Y}}_t)$ can help to capture the overall class-wise distribution on the target domain, synthesizing $(\tilde{\mathcal{X}}_{s2t|\mathcal{Y}_{s}}, \mathcal{Y}_s)$ from $\mathcal{Y}_s$ becomes possible since the label space is shared between clear and adverse weathers. Moreover, another advantage is that source domain data are never touched to train the diffusion model, meaning no source-specific features will be introduced to update the model. In theory, the generated images from such a diffusion model will purely follow the target domain distribution. Therefore, we take the pre-trained UDA model $\mathcal{M}$ from ~\cite{hoyer2023mic}, and prepare a set of predicted labels $\tilde{\mathcal{Y}}_t = \mathcal{M}(\mathcal{X}_t)$, which are referred to as target prior.
\\
{\bf Condition Fusion of Semantic and Structure}
We build our UDAControlNet on top of ControlNet~\cite{zhang2023adding}. As mentioned above, we take target domain images and the target prior, \textit{i.e.,} $(\mathcal{X}_t, \tilde{\mathcal{Y}}_t)$ to train our DM, which is the prerequisite for any other technique in our ControlUDA framework. However, unlike many popular datasets~\cite{everingham2010pascal,lin2014microsoft,deng2009imagenet}, the complex scene structure and object appearances of autonomous driving data, as well as the adverse weathers make it quite challenging to generate high quality data from the noisy $\tilde{\mathcal{Y}}_t$ alone. Hence, other than preparing $\tilde{\mathcal{Y}}_t$, we also utilize pretrained HED~\cite{xie2015holistically} to prepare a set of sketches ${H}_t = \mathcal{H}(\mathcal{X}_t)$, which are less affected across weather variations and appear similarly, as an extra domain-agnostic structure guidance to our DM. Considering that the training of our UDAControlNet is task-specific, we want to prioritize the input conditions on the semantic modality rather than the structure. Therefore, we introduce a residual condition fusion (RCF) module to handle different input modalities at the same time. Specifically, in each training iteration, $\tilde{{y}}_t$ and ${h}_t$ (where ${h}_t \in {H}_t)$ are received by two separate encoders $\mathcal{E}^{seg}$ and $\mathcal{E}^{str}$ to produce ${c}^{seg}_{t}$ and ${c}^{str}_{t}$, and an attention module ${\mathcal{A}}$ is then applied to ${c}^{str}_{t}$ to remove non-salient noisy artifacts but enhance the important salient structures. Afterwards, an early condition fusion is conducted via element-wise addition between the output structural feature and ${c}^{seg}_{t}$, which is then processed by a 1x1 convolution layer. To keep the semantic modality dominant and also make it possible to control how much information is needed from the early fusion, a skip connection is applied from ${c}^{seg}_{t}$ to the early fused conditions, obtaining ${c}^{f}_{t}$. This process is described as,
\begin{align}
    \label{eq:cond_fuse}
    &{c_{t}^{f}= c_{t}^{seg}\oplus\mathcal{K}(c_{t}^{str}\odot({I}\oplus\mathcal{A}(c_{t}^{str}))\oplus c_{t}^{seg})}
\end{align}
where ${I}$ stands for identity matrix, $\oplus$ and $\odot$ are element-wise addition and multiplication operations.\\
{\bf Multi-scale Training}
Due to the existence of small objects that are far away from the camera and their degraded visibility in adverse driving scenes, ControlNet is observed to struggle in handling those small objects. Therefore, we want our network to be capable of generating images at different input scales. For instance, from a raw input condition, we first get its low resolution version ${y}^{ls}_{t}$ by resizing and a high resolution random crop ${y}^{hs}_{t}$ from the raw input, then we include them both into the mini-batch which is represented by ${y}_{t}$. In other words, in each iteration, we want our DM to learn from different input scales at the same time, encouraging it to treat both local and global features equally well. Likewise, we prepare the input structure condition ${h}_{t}$ as well as ${x}_{t}$ which will be used to generate prompt.
\\
{\bf Label-guided Prompt Enhancement} Even though Blip~\cite{li2022blip}, the prompt generator adopted by ControlNet, can provide captions for many natural images to improve the DM training quality, we still find that it can produce inaccurate captions for driving images in adverse weathers. This happens either by mentioning objects that do not exist or sometimes giving only a few vague words to describe an image. We argue that introducing noisy information as such can harm the training quality and make the generated output less aligned with the input conditions. Considering that Blip is a frozen tool and fine-tuning is thus impractical, one feasible solution is to make Blip less influential during prompt generation. Therefore, we propose to map ${y}_{t}$ class-wisely into extra prompt to provide more semantic guidance in the description. Moreover, to make the prompting more weather controllable, we inject target sub-domain {(i)} by name (e.g., night, foggy) to the default Blip prompt. So far, our enhanced prompt, denoted as ${c}^{p}_{(i)}$, is formatted as `\{Sub-domain\} + Blip prompt + \{Label-guidance\}.' (See Fig.\ref{fig:diagram} for an example) . In this way, the resulting prompt's reliance on Blip output is reduced and its correlation with the input semantic condition is built. Following~\cite{zhang2023adding,rombach2022high}, our enhanced prompt is then processed by Clip~\cite{radford2021learning}. Inspired by~\cite{zhang2023adding}, we also perform a dropout~\cite{srivastava2014dropout} to ${c}^{p}_{(i)}$ with a low probability during training to further encourage our DM to learn from the input conditions. Our enhanced prompt makes the DM training more efficient for data under adverse weathers. \\
{\bf Diffusion Training Objective}
Given a ${z}_{{t}^{(i)}}^{0}$ of target sub-domain $(i)$ encoded by $\mathcal{E}$ (pre-trained by SD), the diffusion algorithm
progressively create a noisified
${z}_{{t}^{(i)}}^{\tau}$, with ${\tau}$ indicating the uniformly sampled time step from $\{0,...,T\}$. With L2-norm as its objective, UDAControlNet learns to predict the added Gaussian noise given the time step $\tau$, the enhanced prompt ${c}_{(i)}^{p}$ as well as our fused input conditions ${c}_{t}^{f}$. With $\epsilon^{\tau}\sim\mathcal{N}(\mu^\tau,{\sigma^\tau}^{2})$, we have, 
\begin{align}
    \label{eq:UDAControlNet}
    &{\mathcal{L}_{Diff}=\mathbb{E}_{{z}_{{t}^{(i)}}^{0},\tau,\epsilon,{c}_{(i)}^{p},{c}_{t}^{f}}\|\epsilon^{\tau}-\epsilon_{\phi,\theta}({z}_{{t}^{(i)}}^{\tau},\tau,{c}_{(i)}^{p},{c}_{t}^{f})\|_{2}^{2}}
\end{align}

\subsection{Pseudo Target Data Generation}
\label{sec:inference}
{\bf Pseudo Target Data from Source Labels}
Fueled by its rich hidden knowledge, tuning SD with our proposed UDAControlNet facilitates the model to produce conditioned outputs with high-fidelity. In addition, thanks to the adoption of target prior $\tilde{\mathcal{Y}}_t$, the model can be trained to only resemble target domain distribution. However, since a common label space is shared by the source and target domains, synthesizing pseudo target images from the more accurate source labels thus becomes feasible, especially with extra guidance of structure information that is less sensitive across domains (See Fig.\ref{fig:diagram}{b}). Therefore, from source input conditions, we apply DDIM sampling~\cite{song2020denoising} to the trained UDAControlNet to acquire pseudo target images,
\begin{align}
    \label{eq:cond_fuse}
    &{\tilde{{x}}_{{s2t}^{(j)}}= \mathcal{D}(DDIM_{UDAControlNet}({z}^{T},{c}_{(j)}^{p},{c}_{s}^{f}))}
\end{align}
where $z^{T}\sim\mathcal{N}(0,1)$ and the time step ${\tau}$ gradually reduces from ${T}$ to ${1}$ throughout the diffusion denoising process. $\mathcal{D}$ is the decoder adopted from SD~\cite{rombach2022high} to project the resulting latent representation into pixel space. By randomly specifying the target sub-domain $(j)$ in our enhanced prompt, UDAControlNet produces controlled high-fidelity outputs under various adverse weathers such as snowy, rainy, night, foggy. Thus, we can obtain a dataset $(\tilde{\mathcal{X}}_{s2t|\mathcal{Y}_{s}}, \mathcal{Y}_s)$ as mentioned above, with which we refine the UDA segmentor $\mathcal{M}$ in later stage. (To emphasize the main input condition, here we use $\tilde{\mathcal{X}}_{s2t|\mathcal{Y}_{s}}$ instead of $\tilde{\mathcal{X}}_{s2t}$)\\
{\bf Pseudo Target Data from Target Prior}
Since the training of UDAControlNet is performed on the target domain, naturally we are able to increase the data diversity of target domain by also generating pseudo data conditioned on the target prior. Likewise, we have,
\begin{align}
    \label{eq:cond_fuse}
    &{\tilde{{x}}_{{t}^{(j)}}= \mathcal{D}(DDIM_{UDAControlNet}({z}^{T},{c}_{(j)}^{p},{c}_{t}^{f}))}
\end{align}
As there is no perfect GT label for the target domain, here we only prepare a set of the pseudo target images $\tilde{\mathcal{X}}_{t|\tilde{\mathcal{Y}}_t}$ which will be used in the next phase. We also observe that in the initial training phase (ca. 2 epochs) of UDAControlNet, the model begins to look for a trade-off between keeping the versatile generative power of the original SD and fitting itself to the target distribution, we find that data generated from such model state later helps to generalize UDA segmentor well to other unseen data. Therefore, we also augment the target domain dataset using the initial UDAControlNet checkpoint. Finally we get $\mathcal{X}_{t} = \mathcal{X}_{t}\cup\tilde{\mathcal{X}}^{final}_{t|\tilde{\mathcal{Y}}_t}\cup\tilde{\mathcal{X}}^{init}_{t|\tilde{\mathcal{Y}}_t}.$\\
\subsection{Domain Adaptation}
\label{sec:task}
After the preparation of pseudo target data, the last step is to improve domain adaptation. We argue that further tuning the state-of-the-art UDA segmentor $\mathcal{M}$ using our generated data can further raise the model performance towards the upper limit of domain adaptive semantic segmentation in adverse weathers. We take MIC~\cite{hoyer2023mic} as our baseline and adapt its training scheme for refinement. On top of MIC base objective function, we additionally calculate a selective supervised loss on data generated from source GT labels. As UDA training does not need to differentiate sub-target domains, notations are omitted here. To be specific, we augment the original target set $\mathcal{X}_{t}$ with our $\tilde{\mathcal{X}}_{t|\tilde{\mathcal{Y}}_t}$, enriching the diversity of the target domain space, which not only aims to facilitate the UDA training but also improve model generalization to unseen data. During training we first calculate the baseline loss $\mathcal{L}^{base}$ following ~\cite{hoyer2023mic}. Notably, for self-training on the augmented target domain dataset, new pseudo labels $\Hat{y}_{t}$ are computed on-the-fly instead of still relying on the target prior $\tilde{y}_{t}$.\\
In terms of ($\tilde{\mathcal{X}}_{s2t|\color{orange}{\mathcal{Y}_{s}}}, \color{orange}{\mathcal{Y}_s}$), a question arises: Which label pixels from source GT $\color{orange}{\mathcal{Y}_s}$ can be taken to supervise the output $\tilde{\mathcal{X}}_{s2t|\color{orange}{\mathcal{Y}_{s}}}$? After tuning DM with target prior, our ControlUDA framework is able to build a close loop from source label to pseudo target data back to the source label, \textit{i.e.}, ${\color{orange}{\mathcal{Y}_s}}\xrightarrow{DM}\; \tilde{\mathcal{X}}_{s2t|{\color{orange}{\mathcal{Y}_{s}}}} \xrightarrow{UDA}\;\color{orange}{\mathcal{Y}_s}$. In this way, we turn a seemingly unsolvable chicken and egg problem into one that can be solved via label supervision. However, the training of UDAControlNet with target prior $\tilde{\mathcal{Y}_{t}}$ is noisy, the generated data is therefore not guaranteed to have pixel-wise correspondence with the GT label even though we use the more fine-grained and accurate $\color{orange}{\mathcal{Y}_{s}}$ for inference. Therefore, we need a mechanism to determine which source GT label pixels can be used for their target pseudo data. This involves two cases: (\textbf{\textit{i}}). the predicted label $y^{pred} = \operatorname*{arg\,max}(\mathcal{M}(\tilde{x}_{s2t}))$ agrees with the GT label $\color{orange}{y_{s}}$; (\textbf{\textit{ii}}). $y^{pred}$ differs from $\color{orange}{y_{s}}$.\\
For case (\textbf{\textit{i}}), a double consistency check is conducted by the above mentioned close loop, which means if the pseudo target image generated from $\color{orange}{y_{s}}$ by our DM can be mapped back to the same semantic label through $\mathcal{M}$, it would imply that this mapping is meaningful and can be further encouraged. For case (\textbf{\textit{ii}}), there should be a trade-off between relying on $\color{orange}{y_{s}}$ or not, \textit{e.g.,} using a threshold to differentiate misclassification and false generation. If $y^{pred}$ differs from $\color{orange}{y_{s}}$ above a confidence threshold $\lambda$, we hypothesize that image generation in those pixel regions do not reflect the label condition well, meaning false generation, we therefore do not compute segmentation loss there. In contrast, if $y^{pred}$ differs from $\color{orange}{y_{s}}$ with confidence below $\lambda$, given that our image generation is additionally supported by structure $h_{s}$, we can assume that this is likely to be caused by the misclassification of the undertrained $\mathcal{M}$, therefore, we adopt $\color{orange}{y_{s}}$ for supervision. To summarize, the updated GT adopted to supervise the pseudo target data can be represented as ${\color{orange}{\Hat{{y}}_{s}}} =  {\color{orange}{y_{s}}}^{(y^{pred}={\color{orange}{y_{s}}})} \cup {\color{orange}{y_{s}}}^{{(y^{pred}\neq {\color{orange}{y_{s}}})}\&\mathcal{M}(\tilde{x}_{s2t})<\lambda}$. As last, the supervised cross-entropy segmentation loss on $\tilde{x}_{s2t}$ is written as,
\begin{align}
    \label{eq:source_seg}
    &{\Hat{\mathcal{L}}_{s2t}^{seg}} =  -\mathbb{E}_{\tilde{x}_{s2t}, \color{orange}{\Hat{{y}}_{s}}} \left[{\color{orange}{\Hat{{y}}_{s}}} \log(\mathcal{M}(\tilde{x}_{s2t}))\right]_{(h,w,c)}
\end{align}
Therefore, the total loss of UDA is $\mathcal{L}^{UDA} = \mathcal{L}^{base}+\Hat{\mathcal{L}}_{s2t}^{seg}$.
\begin{figure*}[t!]
\centering
\includegraphics[width=2.0\columnwidth,clip=true,trim=0 0 0 0]{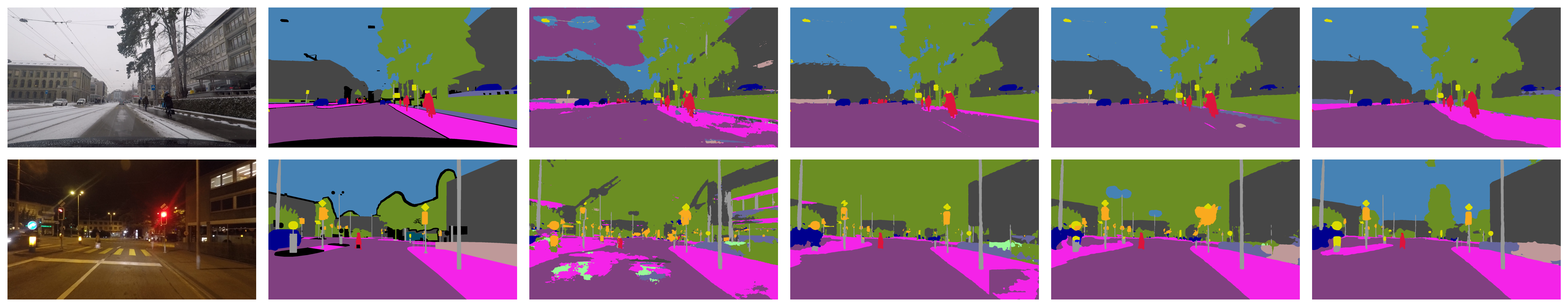}
\caption{{\bf Qualitative comparison of Cityscapes-to-ACDC adaptation on ACDC val set}. Columns from left to right are: target domain inputs; ground-truths; segmentation predictions from DAFormer~\cite{hoyer2022daformer}, HRDA~\cite{hoyer2022hrda}, MIC~\cite{hoyer2023mic} and ControlUDA (ours).}
\label{fig:seg_output}
\end{figure*}
\begin{table*}[htb!]
    \caption{{\bf Cityscapes-to-ACDC adaptation results evaluated on the test set}. We compare the performance of ControlUDA with state-of-the-art methods. Bold indicates the {\bf best} and underline the \underline{second best}. $^{\star}$ means training in weakly supervised setting using extra clear reference images. Regarding backbone architectures for segmentation: `$R$' stands for {\tt ResNet}~\cite{he2016deep}, `$M$' stands for MiT-B5, a transformer-based architecture used in Segformer~\cite{xie2021segformer}. \textit{Among the results officially stated in the papers, MIC was the SOTA method in all.}}
    \centering
    \small
    \setlength{\tabcolsep}{3.5pt}
    \fontsize{7}{11}\selectfont
    \begin{tabular}{c|c|ccccccccccccccccccc|c}
    \hline Method & Arch.. & \rotatebox{90}{road} & \rotatebox{90}{sdwk} & \rotatebox{90}{bldng} & \rotatebox{90}{wall} & \rotatebox{90}{fence} & \rotatebox{90}{pole} & \rotatebox{90}{light} & \rotatebox{90}{sign} & \rotatebox{90}{veg} & \rotatebox{90}{trrn} & \rotatebox{90}{sky} & \rotatebox{90}{psn} & \rotatebox{90}{rider} & \rotatebox{90}{car} & \rotatebox{90}{truck} & \rotatebox{90}{bus} & \rotatebox{90}{train} & \rotatebox{90}{moto} & \rotatebox{90}{bike} & mIoU \\ 
    \hline
    \\[-1.6em]
    MGCDA$^{\star}$~\cite{sakaridis2020map} &$R$&  73.4&  28.7& 69.9& 19.3& 26.3& 36.8& 53.0& 53.3& 75.4& 32.0& 84.6&51.0& 26.1& 77.6&  43.2&  45.9& 53.9& 32.7& 41.5 & 48.7\\
    \\[-1.8em]
    DANNet$^{\star}$~\cite{wu2021dannet} &$R$&  84.3&  54.2& 77.6& 38.0&30.0& 18.9& 41.6& 35.2& 71.3& 39.4& 86.6& 48.7& 29.2& 76.2&  41.6&  43.0& 58.6& 32.6& 43.9& 50.0\\
    \\[-1.8em]
   DAFormer~\cite{hoyer2022daformer} &$M$&  58.4&  51.3& 84.0& 42.7&35.1& 50.7& 30.0& 57.0& 74.8& 52.8& 51.3& 58.3& 32.6& 82.7& 58.3&  54.9&  82.4& 44.1& 50.7& 55.4\\
    \\[-1.8em]
   Refign$^{\star}$~\cite{bruggemann2023refign} &$M$&  89.5&  63.4& 87.3& 43.6&34.3& 52.3& \underline{63.2}& 61.4& {\bf86.9}& 58.5& {\bf95.7}& 62.1& 39.3& 84.1& 65.7&  71.3&  85.4& 47.9& 52.8& 65.5\\
    \\[-1.8em]
   HRDA~\cite{hoyer2022hrda} &$M$&  88.3&  57.9& 88.1& \underline{55.2}&36.7& 56.3& 62.9& 65.3& 74.2& 57.7& 85.9& 68.8& 45.7& 88.5& {\bf76.4}&  82.4&  87.7& 52.7& \underline{60.4}& 68.0\\
   \\[-1.8em]
   MIC~\cite{hoyer2023mic} &$M$&  \underline{90.8}&  \underline{67.1}& \underline{89.2}& 54.5&\underline{40.5}& \underline{57.2}& 62.0& \underline{68.4}& 76.3& {\bf61.8}& 87.0& {\bf71.3}& \underline{49.4}& {\bf89.7}& \underline{75.7}&  {\bf86.8}& \underline{89.1}& {\bf56.9}& {\bf63.0}& \underline{70.4}\\
    \\[-1.7em]
    \hline
    ControlUDA $(\mathrm{ours})$ &$M$&  {\bf94.6}&  {\bf78.0}& {\bf90.3}& {\bf59.0}&  {\bf42.5}& {\bf59.6}& {\bf69.8}& {\bf69.2}& \underline{81.3}& \underline{61.0}& \underline{90.3}& \underline{70.1}& {\bf50.8}& \underline{89.0}& 75.5& \underline{85.8} &{\bf89.6} & \underline{53.3}& 57.3&{\bf72.0}\\ 
    \hline
\end{tabular}
\label{tab:citytoacdc}
\end{table*}

\section{Experiments}
We experimentally demonstrate the competitiveness and superiority of ControlUDA to boost domain adaptive semantic segmentation in adverse conditions. We report our leading UDA model performance on popular benchmark datasets. Extensive experiments and analysis are also conducted to verify our design. Details about our implementation are provided in the Supplementary. \\

\subsection{Benchmark Datasets}
\noindent\textbf{Cityscapes}~\cite{cordts2016cityscapes} is adopted as the labelled clear weather source domain, containing 2,975 19-categorical urban scene images with pixel-wise annotation. The original image resolution is 2048×1024.\\
\noindent\textbf{ACDC}~\cite{sakaridis2021acdc} is considered as our unlabelled target domain under adverse weather conditions such as fog, snow, rain, as well as nighttime. It contains 4,006 images of 1920×1080 resolution, including 1,600 training images, 406 validation images, and 2000 test images. However, the ground-truth of the test set is not publicly available, evaluation results on which can be attained by online submission.\\

\subsection{Implementation Details}
\label{sec:implementation_details}
We implement ControlUDA on NVIDIA Quadro RTX 8000 with 48 GB memory. In terms of the offline preparation of target prior $\tilde{\mathcal{Y}}$ from the pre-trained UDA segmentor $\mathcal{M}$, as well as sketch from the pre-trained edge detector HED~\cite{xie2015holistically}, we take the full resolution RGB images from ACDC and Dark Zurich datasets $1920$×$1080$. This also applies to obtaining HED sketches for source domain cityscapes images, where the processed resolution is $2048$×$1024$.\\
Regarding the training of UDAControlNet, we build our implementation on top of ControlNet~\cite{zhang2023adding}. We use AdamW~\cite{loshchilov2017decoupled} optimizer with a learning rate of $1\times 10^{-5}$, and this learning rate is applied to ControlNet $\phi$ related modules. Additionally, following the suggestions of ControlNet, we unlock the decoder of SD ($\theta$) but with a relatively smaller learning rate $5\times 10^{-6}$, which is claimed to be able to obtain better visual quality. For the multi-scale training, considering the fact that ControlNet only support inputs with resolution being multiplier of 64, and in the meanwhile, the aspect ratio should be the same as the raw images from the target domain to prevent distortion, therefore, the largest reasonable training resolution is $1344$×$768$. In other words, we have a resized version of the raw image into $1344$×$768$ resolution and a $1344$×$768$ random crop from the raw image. Due to the GPU memory limitation, we set the batch size to $4$ and apply gradient accumulation~\cite{gradaccu} with a period of $4$, which is supposed to match an equivalent batch size of $16$. To enable our multiscale training, we change the default Blip prompt usage setting of ControlNet from offline prompt generation into online prompt generation. In terms of prompt dropout probability, considering that the weather control after training need be managed by the prompt, we only set the dropout probability to $0.01$ Training a UDAControlNet takes $160$ hours on a single GPU.\\ 
When doing inference to our trained UDAControlNet model, if the input conditions come from the source clear weather domain dataset (e.g., Cityscapes), we take the raw images of resolution $2048$×$1024$ and only pass them to Blip encoder to generate prompt, such that no source-specific (domain related) information is introduced to infer our UDAControlNet. Additionally, since the ground-truth labels $y_{s}$ from the source domain Cityscapes dataset contain the so called `don't care' regions marked as class $255$ in black color, making their direct adoption as input label condition impractical due to this format conflict. Therefore, the original ground-truth labels are only adopted to acquire label-guided prompts as the `don't care' regions are easy to ignore in prompts. However, to prepare input label condition to $\mathcal{E}^{seg}$, we follow~\cite{shen2023loopda} and perform label fusion, merging $y_{s}$ with the label predicted by $\mathcal{M}$. Therefore, the enhanced prompt ${c}^{p}_{(i)}$ consists of Blip prompt injected by the target sub-domain name {(i)} and the label-guided prompt from $y_{s}$. By randomly specifying the sub-domain name, UDAControlNet is supposed to produce an output aligning with the mentioned weather.
Afterwards, $\mathcal{E}^{seg}$ receives one-hot input label condition with $K$ channels ($K$ is the number of semantic classes). However, when generating $\tilde{\mathcal{X}}_{t|\tilde{\mathcal{Y}}_{t}}$ instead of $\tilde{\mathcal{X}}_{s2t|\mathcal{Y}_{s}}$, nothing has to be modified to the target prior $\tilde{\mathcal{Y}}_{t}$. Inference the a whole dataset takes $24$ hours on $5$ GPUs.\\
For the task refinement training of UDA segmentation, in our main paper, we take MIC~\cite{hoyer2023mic} as our baseline segmentor. When training domain adaptation, we follow~\cite{hoyer2023mic} and use AdamW~\cite{loshchilov2017decoupled} optimizer with a learning rate of $6\times 10^{-5}$ following~\cite{hoyer2022daformer,hoyer2022hrda,hoyer2023mic}.\\

\begin{table}[t!]
    \caption{{\bf Performance comparison on the val set} of Cityscapes-to-ACDC adaptation benchmark. $^{\star}$ means weak supervision with reference data.}
    \centering
    \setlength{\belowcaptionskip}{-12pt}
    \setlength{\tabcolsep}{3pt}
    \fontsize{8}{12}\selectfont
    \begin{tabular}{c|cccc}
        \hline
         Method              &  HRDA~\cite{hoyer2022hrda}   &     Refign$^{\star}$~\cite{bruggemann2023refign}   &      MIC~\cite{hoyer2023mic}   &     ControlUDA $(\mathrm{ours})$                      \\ \hline
        mIoU    & 65.3    & 65.4 & 69.6      & {\bf 71,8}  
        \\[-1.8em]
       \\ \hline
    \end{tabular}
   \label{tab:val_city2acdcdz}
\end{table}

\begin{table}[htb!]
    \caption{ {\bf Quantitative comparison of image generation quality} among popular generative models. Bold stands for the {\bf best}.}
    \centering
    \small
    \setlength{\tabcolsep}{7pt}
    \fontsize{7}{11}\selectfont
    \begin{tabular}{c|ccc}
        \hline Method &  FID$\downarrow$~\cite{heusel2017gans} & LPIPS$\downarrow$~\cite{zhang2018unreasonable} & MS-SSIM$\uparrow$~\cite{wang2003multiscale}  \\ 
        \hline
        \\[-1.6em]
        OASIS~\cite{hoyer2022hrda}  &  163.12& 0.68 & 0.50 \\
        \\[-1.8em]
        ControlNet~\cite{bruggemann2023refign} &  99.12& 0.70 & 0.49 \\
        \\[-1.6em]
        \hline
        UDAControlNet$(\mathrm{ours})$ &  {\bf 91.34}&  {\bf 0.59}& {\bf 0.60}\\ 
        \hline
    \end{tabular}
\label{tab:metric_gen}
\end{table}
\begin{figure}[t!]
    \centering
    \setlength{\belowcaptionskip}{-1pt}
    \includegraphics[width=\linewidth]{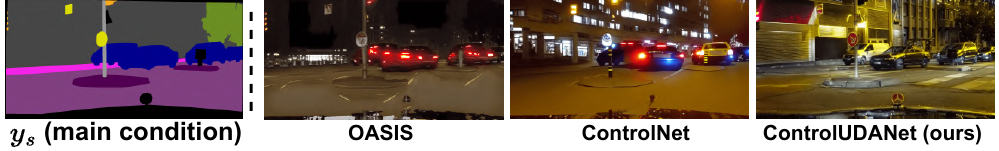}
    \caption{{\bf Visual comparison of different generative models} Given the same input semantic condition, outputs from different approaches are visualized. For OASIS, due to its limitation, we sample till the desired weather appears.}
    \label{fig:gen_img_compare}
\end{figure}
\subsection{Model Evaluation}
{\bf Benchmark Evaluation} We evaluate ControlUDA and compare it with SOTA methods for domain adaptive semantic segmentation in adverse conditions. As shown in Table~\ref{tab:citytoacdc}, ControlUDA shows superior performance to the SOTA methods on test set of Cityscapes-to-ACDC adaptation, pushing the upper bound of this unsupervised task further and obtaining $71.6$ mIoU even without any weak supervision from reference data. An visual impression is given in Fig.\ref{fig:seg_output}. The superiority of ControlUDA is also verified on the val sets of the benchmarks (shown in Table~\ref{tab:val_city2acdcdz}), and it achieves $71.8$ and $47.6$ mIoU respectively, outperforming SOTA methods by considerable margins.\\
\noindent{\bf Image Quality Metrics} To show the quality of our data generation, in Table~\ref{tab:metric_gen} we compare with OASIS~\cite{sushko2020you} (a popular GAN-based method) and ControlNet~\cite{zhang2023adding} (the strongest diffusion baseline) on the widely adopted metrics~\cite{heusel2017gans,zhang2018unreasonable,wang2003multiscale}. FID~\cite{heusel2017gans} has proven to correlate with human preference on image fidelity and diversity when fake and real datasets are compared. LPIPS~\cite{zhang2018unreasonable} measures the perceptual quality of fake data compared to real data via deep feature space. MS-SSIM~\cite{wang2003multiscale} is a statistical metric to calculate multi-scale structural similarity between two images. 
We randomly pick $30$ labels from $\mathcal{Y}_{s}$ and $\tilde{\mathcal{Y}_{t}}$ respectively and generate 10 images per label. For LPIPS and MS-SSIM, we compare the generated images with the corresponding real image and the score is averaged on the 600 image pairs for each model. In terms of FID, for each method we compare the $600$ images with real full target dataset. Note that the compared methods are trained under their default setting without modification, and crop size is $1344$×$768$ for all methods.
We find that UDAControlNet outperforms other methods in all metrics when handling data generation in these challenging adverse conditions. An example is shown in Fig.\ref{fig:gen_img_compare}, where all methods take the same label condition. We observe that, not only appearing with lower fidelity, compared to ours, other methods also struggle when vehicles are overlapped in the semantic label.\\
To further show the superiority of UDAControlNet to other generative methods for data synthesis, in Table~\ref{tab:table_source_gen}, we train the segmentor merely on $(\mathcal{X}_{s}, \mathcal{Y}_{s})$ and also $(\tilde{\mathcal{X}}_{s2t|\mathcal{Y}_{s}}, \mathcal{Y}_{s})$ with our thresholding, but we compare the model performance on target val set. We observe that using data generated by UDAControlNet assists the baseline segmentor to obtain a remarkable gain (+$9.26$ mIoU), achieving 65.05 mIoU and outperforming ControlNet by $5.67$ mIoU. This indicates that UDAControlNet is more suitable for this UDA task in adverse conditions.  

\begin{table}[t!]
    \centering
    \setlength{\belowcaptionskip}{-1pt}
    \setlength{\tabcolsep}{1.5pt}
    \fontsize{8}{12}\selectfont
    \begin{tabular}{c|c|c|c|c} \hline 
        \multirow{2}{*}{Method}    &   \multicolumn{4}{c} {Train on $(\mathcal{X}_{s}, \mathcal{Y}_s)$ + $(\tilde{\mathcal{X}}_{s2t|\mathcal{Y}_{s}}, \mathcal{Y}_s)$} \\ \cline{2-5}
                &   Baseline$^{\dagger}$    &   OASIS~\cite{sushko2020you}    &   ControlNet~\cite{zhang2023adding}    &   UDAControlNet (Ours)    \\ \hline\hline
        mIoU    &   {55.79}      &   {53.38}      &   {59.38}      &   {\bf 65.05} \\\hline
    \end{tabular}
    \caption{{\bf Segmentation performance comparison on ACDC val set using pseudo data generated from source labels.} For comparison, pseudo target data are generated by different methods to train the segmentor. $^{\dagger}$ means no generated pseudo data are used for training. Note that models are trained without target data ${\mathcal{X}_{t}}$.}
    \label{tab:table_source_gen}
\end{table}

\subsection{Ablative Analysis}
{\bf Ablation on pseudo target data}
We ablate on our three types of generated pseudo target data for tuning $\mathcal{M}$. Comparing Base1 and Base2 in Table~\ref{tab:ablation_gen_type}, we observe that tuning $\mathcal{M}$ without our generated pseudo data brings almost no change. Comparing from ablation $(i)$ to $(iv)$, we find that each type of pseudo data has its contribution to the final performance gain, and combining three of them yields the best performance on the target val set ($71.8$ mIoU). Among all types of target pseudo data, owing to the existence of source ground-truth labels, training the segmentor with $\tilde{\mathcal{X}}_{s2t|\mathcal{Y}_{s}}$ brings the largest contribution to the performance gain (see row $(iii)$).
\begin{figure}[t!]
    \centering
    \setlength{\belowcaptionskip}{-5pt}
    \includegraphics[width=\linewidth]{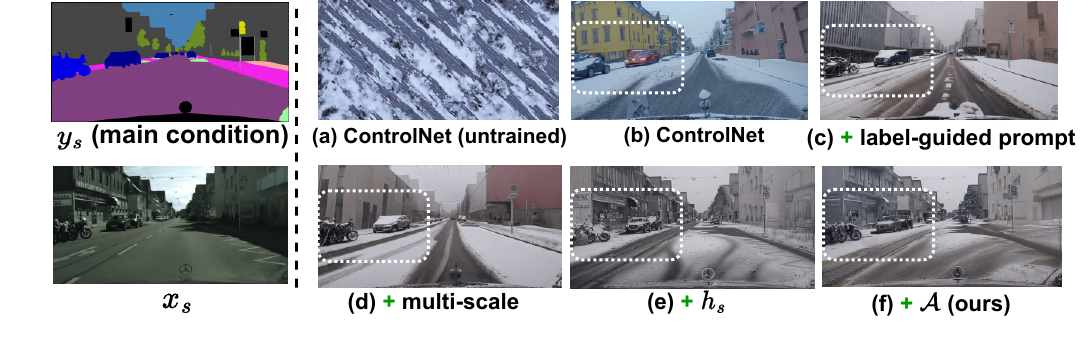}
    \caption{{\bf Visual ablation of UDAControlNet component}. Given the same input semantic condition, we train UDAControlNet with different configurations. From (a) to (f), components are added on top of the previous one.}
    \label{fig:ablation_component}
\end{figure}\\
\begin{table}[t!]
    \centering
    \setlength{\tabcolsep}{4pt}
    \fontsize{8}{12}\selectfont
    \begin{tabular}{c|cc|ccc|cc}
        Method              &   $\mathcal{X}_{s}$  &   $\mathcal{X}_{t}$   &    {\bf +}$\tilde{\mathcal{X}}_{s2t|\mathcal{Y}_{s}}$   &      {\bf +}$\tilde{\mathcal{X}}^{final}_{t|\tilde{\mathcal{Y}}_t}$  &      {\bf +}$\tilde{\mathcal{X}}^{init}_{t|\tilde{\mathcal{Y}}_t}$      &   mIoU  &   $\Delta$                  \\ \hline
        Base1~\cite{hoyer2023mic}         &  &       &           &           &                     &   69.6  &   \textcolor{green}{+}0.0   \\
        \\[-2.0em]
        Base2~\cite{hoyer2023mic}        & \cmk  &   \cmk     &           &           &                 &   69.8  &   \textcolor{green}{+}0.2   \\ \hline
        (\romannum{1})       & \cmk  & \cmk  &  &      &      \cmk          &    70.1  &   \textcolor{green}{+}0.5  \\ \\[-2.0em]
        (\romannum{2})      & \cmk  & \cmk  &  & \cmk   &                 &  70.3  &   \textcolor{green}{+}0.7  \\
        \\[-2.0em]
        (\romannum{3})      & \cmk  & \cmk  &  \cmk &      &                &   70.8  &   \textcolor{green}{+}1.2   \\
        \\[-2.0em]
        (\romannum{4})       & \cmk  & \cmk  & \cmk &     \cmk  &                &    71.3  &   \textcolor{green}{+}1.7  \\ \\[-2.0em]
        (\romannum{5})      & \cmk  & \cmk  & \cmk &  &        \cmk           &  71.3  &   \textcolor{green}{+}1.7  \\
        \\[-2.0em]
        (\romannum{6})      & \cmk  & \cmk  & \cmk & \cmk  & \cmk &  {\bf 71.8}  &   \textcolor{green}{+}2.2  \\ \hline
    \end{tabular}
    \caption{{\bf Ablation on the input data types adopted to train the UDA segmentor $\mathcal{M}$ }. We take MIC as the baseline, where the first one means pre-trained checkpoint, and the second baseline means tuning $\mathcal{M}$ without our generated data. Evaluated on ACDC val.} 
    \label{tab:ablation_gen_type}
\end{table}
\noindent{\bf Ablation on model components} After training with target prior $\tilde{\mathcal{Y}}_{t}$ on different model configurations, we visualize the impact of each UDAControlNet component in Fig.\ref{fig:ablation_component} conditioned on source label $y_{s}$. After tuning SD with default ControlNet, we see change from $(a)$ to $(b)$, where the model is starting to reflect $y_{s}$. However, $(b)$ cannot follow $y_{s}$ well (e.g., motors) due to the imperfect $\tilde{\mathcal{Y}}_{t}$ and lack of sufficient prompting during training. $(c)$ improves this with our enhanced prompt with label guidance, but the details of the overlapped motors and far objects are not properly handled, which is the reason to introduce multi-scale training in $(d)$. However, $(d)$ still struggles to give reasonable generation when there is an overlap of objects in $y_{s}$, for this reason $h_{s}$ is added in $(e)$ to enable structural awareness. This solves model confusion on overlaps, but is affected the noise in $h_{s}$. Therefore, incorporating structural attention using $\mathcal{A}$ in our final UDAControlNet gives the best fidelity output, showing high correspondence with the raw $x_{s}$ even though the model is not trained on it.\\
\noindent{\bf Ablation on thresholds} To understand the impact of $\lambda$, in Table~\ref{tab:threshold}, we make an ablation for threshold values, and find that applying no threshold at all results in the lowest mIoU as the model encourages a supervision signal only when $y^{pred}$ agrees with $y_{s}$, which is not difficult for the network to learn. In addition, applying too large or too small $\lambda$ also provide limited support for tuning $\mathcal{M}$. We achieve SOTA results when setting $\lambda$ to $0.85$, which suggests that misclassification and false generation reach the best trade-off at this point. Setting $0.85$ to $0.65$ also indicates that using more label pixels from $y_{s}$ does not mean better performance.

\begin{table}[t!]
    \centering
    \setlength{\tabcolsep}{5pt}
    \setlength{\belowcaptionskip}{-5pt}
    \fontsize{8}{12}\selectfont
    \begin{tabular}{c|ccccc}
        $\lambda$              &  {no threshold}   &   {$0.65$}   &     {$0.75$}   &       {$0.85$}   &  {$0.95$}                      \\ \hline
        mIoU    & 70.7    & 70.8 & 71.1      & {\bf 71.8}           & 71.1       
        \\[-1.8em]
       \\ \hline
    \end{tabular}
    \caption{ {\bf Ablation on the impact of threshold value} $\lambda$ for refinement training of the UDA segmentor. Evaluated on ACDC val.}
    \label{tab:threshold}
\end{table}

\subsection{Generalizability Analysis}
In Table~\ref{tab:table_dg}, we experimentally demonstrate that our ControlUDA framework not only increase the model adaptability on the target domain, but also facilitates its generalizability on unseen data. When evaluating our segmentor adapted from Cityscapes-to-ACDC on other datasets such as BDD100k~\cite{yu2020bdd100k}, Mapillary~\cite{neuhold2017mapillary}, Dark Zurich~\cite{sakaridis2019guided} and Night Driving~\cite{dai2018dark} val sets, thanks to the versatile knowledge mined from SD, we find our model better generalized on unseen data then the MIC baseline, and it even improves the accuracy on the source Cityscapes val set, attaining 81.37 mIoU. Interestingly, we also observe that UDA training brings a huge boost to the model generalizability compared to a source-only model, although the UDA segmentor is trained in an unsupervised manner.\\
Although we tailor our ControlUDA framework for handling UDA in adverse conditions, we also show that our method can be adopted for other UDA benchmarks and achieve SOTA performances (see Table~\ref{tab:val_gta}). 

\begin{table}[t!]
    \centering
        \setlength{\tabcolsep}{5pt}
        \setlength{\belowcaptionskip}{-3pt}
    \fontsize{8}{12}\selectfont
    \begin{tabular}{cc|cccc} \Xhline{2\arrayrulewidth}
        \multirow{2}{*}{Method}    &    \multicolumn{5}{c}{ Cityscapes as Source Domain} \\ \cline{2-6}
            &  C  &  $\rightarrow$B    &  $\rightarrow$M    &    $\rightarrow$D & $\rightarrow$N      \\ \Xhline{2\arrayrulewidth}
           MIC~\cite{hoyer2023mic} (City-only) &   80.73     &   54.56     &  65.65      &  28.42 &49.29\\
        MIC~\cite{hoyer2023mic} (City-to-ACDC) &   80.74      &   59.01      &  70.57     &   41.00 &53.50\\
        \hline
        Ours (City-to-ACDC)    &   {\bf81.37}      &   {\bf59.29}      &   {\bf70.90}      &   {\bf45.32} &{\bf56.79} \\\hline
        \Xhline{2\arrayrulewidth}
    \end{tabular}
    \caption{{\bf The improved domain generalizability of ControlUDA on unseen data}. C, B, M, D and N denote {\bf C}ityscapes, {\bf B}DD100k, {\bf M}apillary, {\bf D}ark Zurich and {\bf N}ight Driving val set respectively.} 
    \label{tab:table_dg}
\end{table}

\begin{table}[t!]
    \centering
    \small
    \setlength{\tabcolsep}{1.5pt}
    \fontsize{7}{11}\selectfont
    \begin{tabular}{c|cc}
    \hline Method &  GTA5-to-Cityscapes~\cite{richter2016playing} & Synthia-to-Cityscapes~\cite{ros2016synthia}  \\ 
    \hline
    \\[-1.6em]
    DAFormer~\cite{hoyer2022hrda}  & 68.3&  60.9 \\
    \\[-1.8em]
    HRDA~\cite{hoyer2022hrda} & 73.8&  65.8 \\
    \\[-1.8em]
    MIC~\cite{hoyer2023mic} &  75.9& 67.3\\
    \\[-1.8em]
    \hline
    ControlUDA $(\mathrm{ours})$ &  {\bf 76.5}&  {\bf 68.7}\\ 
    \hline
\end{tabular}
\caption{{\bf mIoU comparison on Cityscapes val set} of other benchmarks such as GTA5-to-Cityscapes and Synthia-to-Cityscapes.}
\label{tab:val_gta}
\end{table}


\section{Conclusion}
In this paper, we propose ControlUDA, a framework tailored for UDA segmentation under adverse weather conditions, addressing the challenges faced by existing works based on data generation. To train a DM that performs cross-domain data synthesis for the preparation of pseudo target images, it compensates for missing the target domain ground-truth labels by using target prior knowledge from a pre-trained segmentor to tune the DM. For training the DM, ControlUDA is empowered by UDAControlNet, a condition-fused multi-scale and prompt-enhanced diffusion tuning network particularly aimed at high-fidelity image generation under adverse weather conditions. The resulting target-specific DM is capable of generating fidelity condition-controllable pseudo target data conditioned on source ground-truth labels. Tuning the segmentor on our generated data substantially boosts the UDA performance of semantic segmentation in adverse conditions. 
{
    \small
    \bibliographystyle{ieeenat_fullname}
    \bibliography{main}
}

\end{document}